\documentclass[sigconf]{acmart}

\usepackage{tabularx}
\usepackage{multirow}

\AtBeginDocument{%
  }

\setcopyright{acmlicensed}
\copyrightyear{2025}
\acmYear{2025}
\acmDOI{XXXXXXX.XXXXXXX}
\acmConference {under review}
\acmISBN{-}
\renewcommand\footnotetextcopyrightpermission[1]{}

\acmSubmissionID{2040}
\begin{document}

\title{UniFlowRestore: A General Video Restoration Framework via Flow Matching and Prompt Guidance}

\author{Shuning Sun, Yu Zhang, Chen Wu, Dianjie Lu, Dianjie Lu, Guijuan Zhan, Yang Weng, Zhuoran Zheng}
\authornote{Corresponding author: Zhuoran Zheng.email: zhengzr@njust.edu.cn}

\renewcommand{\shortauthors}{Sun et al.}

\begin{abstract}
  Video imaging is often affected by complex degradations such as blur, noise, and compression artifacts. Traditional restoration methods follow a “single-task single-model” paradigm, resulting in poor generalization and high computational cost, limiting their applicability in real-world scenarios with diverse degradation types. We propose \textbf{UniFlowRestore}, a general video restoration framework that models restoration as a time-continuous evolution under a prompt-guided and physics-informed vector field. A physics-aware backbone PhysicsUNet encodes degradation priors as potential energy, while PromptGenerator produces task-relevant prompts as momentum. These components define a Hamiltonian system whose vector field integrates inertial dynamics, decaying physical gradients, and prompt-based guidance. The system is optimized via a fixed-step ODE solver to achieve efficient and unified restoration across tasks. Experiments show that UniFlowRestore delivers state-of-the-art performance with strong generalization and efficiency. Quantitative results demonstrate that UniFlowRestore achieves state-of-the-art performance, attaining the highest PSNR (33.89 dB) and SSIM (0.97) on the video denoising task, while maintaining top or second-best scores across all evaluated tasks.
\end{abstract}

\ccsdesc[500]{Computing methodologies~Computer vision}
\ccsdesc[300]{Computing methodologies~Reconstruction}
\ccsdesc[300]{Computing methodologies~Machine learning algorithms}

\keywords{Video Restoration, Flow Matching, Prompt Guidance, All-in-one.}

\maketitle

\section{Introduction}

\begin{figure}
    \centering
    \includegraphics[width=1\linewidth]{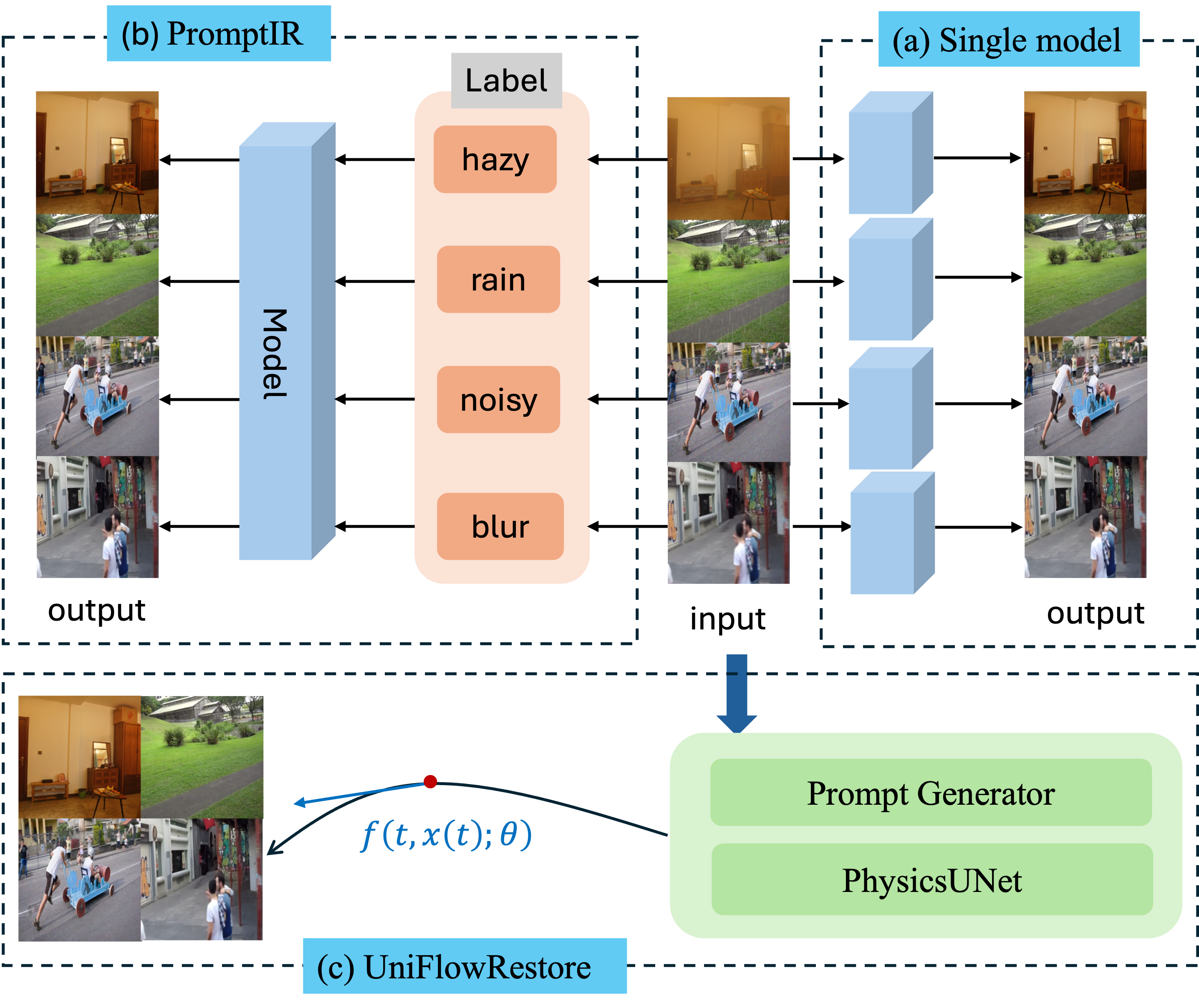}
    \caption{Comparison of different video restoration paradigms. (a) Single-task models train a dedicated network for each degradation type, resulting in high maintenance cost and poor generalization. (b) PromptIR introduces task labels and prompt tuning to enable multi-task restoration, but relies heavily on task annotations and lacks temporal modeling. (c) Our method formulates video restoration as a continuous evolution under a prompt-guided vector field, enabling unified and label-free restoration across multiple degradations with improved temporal coherence.}
    \vspace{-2mm}
    \label{fig:enter-label}
\end{figure}

With the rapid proliferation of video capture devices and content platforms, real-world video data has become increasingly susceptible to a variety of complex degradation factors, such as motion blur, sensor noise, and compression artifacts \cite{vasu2017local, chen2024svt, hyun2013dynamic}. These degradations, jointly induced by both environmental and human factors, not only severely compromise the visual quality of videos but also constrain the performance of subsequent tasks like video analysis, intelligent surveillance, and trajectory prediction. Consequently, there is a pressing need to develop a unified and efficient algorithm for video restoration.

Currently, most traditional video restoration methods follow a fixed “single-task, single-model” paradigm, where a dedicated network is trained for each specific type of degradation, yielding certain advancements \cite{zhang2024blur, ji2022multi, li2024stformer, li2018benchmarking}. However, when confronted with multi-task or multi-degradation scenarios, this paradigm tends to suffer from poor generalization and limited transferability; furthermore, its computational cost grows exponentially as the number of tasks or degradation types increases. In addition, these methods rely heavily on specific degradation priors (for example, in dehazing and underwater image enhancement, only the dark channel prior is typically considered \cite{sathya2015underwater}), rendering them ill-suited to real-world scenarios where multiple forms of degradation may be entangled or even unknown.

In recent years, several studies have explored universal models or multi-task learning frameworks to unify video restoration tasks, striving to handle multiple degradation types within a single model \cite{li2022all, li2020all, nah2021ntire}. However, since these approaches generally lack physical constraints and do not incorporate carefully designed distribution transition pathways, they often struggle to balance model size control and adaptability. Task-agnostic models, due to insufficient fine-grained modeling of specific degradation characteristics, frequently yield suboptimal restoration performance; meanwhile, task-specific models must switch networks or expand parameters whenever a new task arises, leading to substantially increased deployment complexity. Most existing methods still rely on prior knowledge of the degradation type to parameterize the network via meta-learning \cite{fan2019general}.

A currently popular method, PromptIR \cite{potlapalli2023promptir}, offers a general and efficient plug-in module augmented by prompts, enabling the restoration of images suffering from a wide variety of degradation types and severities—even in cases where the exact form of corruption is initially unknown. However, when confronted with video data that exhibit stronger temporal correlations, larger data volumes, and evolving degradation distributions across the temporal dimension, PromptIR’s lack of inter-frame continuity modeling and dynamic information handling becomes a significant limitation. Compared to single-image restoration models, our work explicitly accounts for temporal modeling and optimizes a distribution transition path in the time dimension, providing faster inference speed and stronger generalization for complex, multiform degradation scenarios in videos.

We present UniFlowRestore, a universal framework for video restoration that seamlessly integrates flow-matching modeling with prompt-based conditional guidance. At its core lies PhysicsUNet, a unified and physics-aware network that implicitly encodes a general prior for video degradation. Complementing this, a learnable PromptGenerator dynamically generates task-relevant prompts to guide feature reconstruction, enabling adaptive and task-specific restoration.

We formulate the restoration process as a continuous dynamical evolution governed by a Hamiltonian system, where the image state is represented as position variables and the prompt information as momentum variables. The Hamiltonian function \( H(q, p) = K(p) + U(q) \) defines the system’s total energy, with \( K(p) \) derived from the PromptGenerator and \( U(q) \) from PhysicsUNet. To solve this system efficiently, we discretize the constructed vector field—comprising both physical degradation and prompt-guided terms—using a fixed-step ODE solver. This yields a stable and efficient iterative optimization path, enabling \textbf{UniFlowRestore} to generalize across diverse video restoration tasks and scenarios.
In addition, we collected a large-scale dataset that includes multiple image and video restoration tasks. 
Major contributions of this paper are as follows:
\vspace{-2mm}
\begin{itemize}
\item We propose a unified video restoration framework called UniFlowRestore, which integrates flow matching and a prompt-based guidance mechanism. Without the need to predefine degradation types, this framework automatically adapts to various complex degradation scenarios, achieving high-quality video restoration across multiple tasks and degradation types. This approach breaks from the traditional “single task, single model” paradigm, offering a high degree of task generalization and adaptability to complex degradations.
\item We design a dynamic prompt generation module, PromptGenerator, capable of producing feature-guidance information tailored to different tasks. This guides the network in accurately modeling and restoring diverse degradation types. In addition, we construct a vector-field formulation that merges physical attenuation and prompt-related semantic information, and introduce a fixed-step ODE solver for continuous dynamic optimization. This improves both the stability and efficiency of the restoration process.
\item We build a dataset and conduct extensive experiments on multiple video restoration task datasets. The results show that UniFlowRestore outperforms existing methods on numerous metrics, balancing both restoration performance and computational efficiency, thereby demonstrating strong generality and practicality.
\end{itemize}

\begin{figure*}
    \centering
    \includegraphics[width=\textwidth, height=0.6\textheight]{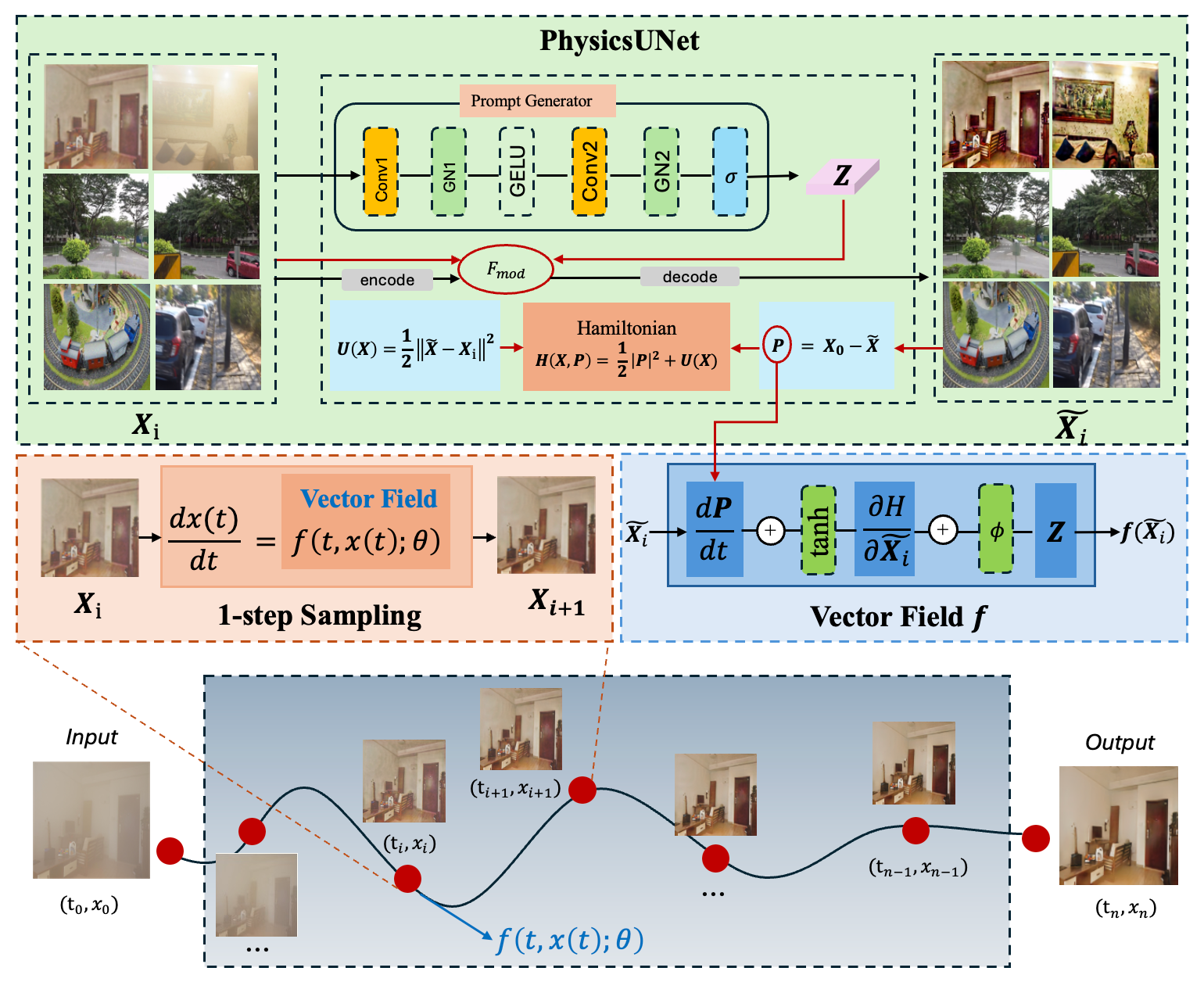}
    \caption{Overview of the UniFlowRestore framework. Given a degraded frame $X_i$, the \textbf{Prompt Generator} extracts a task-aware prompt $Z$ to modulate the PhysicsUNet backbone. The network predicts a clean intermediate frame $\tilde{X}_i$ and constructs the Hamiltonian energy with potential $U(X)$ and momentum $P$. These are injected into a time-continuous vector field $f(X_i)$, which governs the image evolution through fixed-step ODE integration. By sampling along the flow $f(t, x(t); \theta)$, the degraded input is progressively restored toward a clean state.}
    \label{fig:framework}
    \vspace{-3mm}
\end{figure*}

\section{Related Works}
\textbf{All-in-one Restoration.} During video acquisition, compression, and transmission, various types of degradation—such as blurring, noise, compression artifacts, and raindrop occlusion—are likely to occur\cite{zhang2018ffdnet, tsai2022banet, qu2019enhanced, li2017aod}. Traditional methods typically follow a “single-degradation modeling” approach that manually identifies the degradation type and uses a dedicated model for inference\cite{zhang2018density, zhang2018ffdnet, zhang2017learning, zhang2021plug, shen2019human}. However, in real-world scenarios, multiple degradations often co-occur, making it difficult for such methods to adapt well to complex environments, and resulting in limited cross-task transferability. To address this bottleneck, researchers have begun exploring cross-task image (video) restoration techniques, including general-purpose network architectures, multi-task parameter-sharing frameworks, and architecture search–based optimizations \cite{jiang2024survey}. Some existing work introduces task labels or degradation classifiers to guide the model’s behavior when switching between different sub-tasks \cite{li2020all, potlapalli2023promptir}. However, these methods are vulnerable to label misclassification or weak prompt representation, leading to inaccuracies in degradation type recognition.To overcome this challenge, Review Learning has been proposed \cite{su2024review}. It adopts sequential learning and sample reviewing strategies to mitigate information contamination and catastrophic forgetting, thus enhancing the generalization ability and stability of a unified model during multi-task training. Additionally, methods like Task-Aware Modulation dynamically adjust model parameters through a conditional encoder, adapting to diverse types of video degradation 
\cite{cheng2023cross}.However, most of these approaches lack explicit modeling of the degradation evolution process and struggle to capture how temporal degradations propagate across consecutive frames. They also often face challenges in balancing model complexity with task adaptability. Consequently, designing an all-in-one video restoration model with robust temporal modeling, semantic adaptability, and efficient training remains crucial.

\noindent\textbf{Prompt-guided Image/Video Restoration.} In recent years, prompt-based techniques have gradually expanded from the field of natural language processing to computer vision tasks\cite{wang2023multitask, wang2022learning}. The core idea is to inject learnable prompt information into the model input to regulate model behavior and enhance generalization capabilities. In the field of image restoration, prompt-guided mechanisms have been adopted in unified multi-task models to guide the model in distinguishing different degradation scenarios through prompts, enabling conditionally adaptive restoration. Compared with traditional hard-coded or label-based control methods, prompt guidance offers greater flexibility and generalization.
Existing methods, such as PromptIR \cite{potlapalli2023promptir} , ProRes \cite{ma2023prores} , and Prompt-in-Prompt (PIP) \cite{li2023prompt}, insert visual or multimodal prompts into the input image or intermediate features to guide the model in jointly modeling tasks like denoising, deblurring, and artifact removal, achieving remarkable performance.
Although these methods perform well on static image processing, they lack the ability to model inter-frame consistency and dynamic distribution changes in long video sequences, making it difficult to handle complex scenarios where degradation patterns vary over time.

\noindent\textbf{Flow Matching Modeling.} Flow matching has recently gained attention as an efficient continuous modeling approach in the field of generative modeling. This class of methods constructs a deterministic vector field to directly learn the mapping between data distributions, thereby avoiding the cumbersome forward/reverse sampling process typically required in traditional diffusion models \cite{lipman2022flow, gat2024discrete, gao2024flow}. Thanks to its low computational complexity and strong representational capacity, flow matching has been applied in various domains, including image generation \cite{ramesh2022hierarchical, esser2024scaling}, speech synthesis \cite{yun2025flowhigh}, and point cloud upsampling \cite{liu2025efficient}.
In this work, we formulate the process of multi-degradation video restoration as a time-evolving continuous optimization problem. By introducing physical priors and semantic prompt information, and drawing inspiration from Hamiltonian dynamics\cite{deng2025denoising}, we construct a vector field that guides the evolution of degraded frames. Experimental results demonstrate that the physics-guided flow matching mechanism effectively combines the flexibility of neural networks with the rigor of physical systems, providing a generalizable and unified modeling framework for multi-degradation restoration tasks.

\section{Methodology}

\subsection{Overview}

Given a degraded image frame $\mathbf{X}_0 \in \mathbb{R}^{3 \times H \times W}$, we formulate the restoration process as a time-continuous dynamical system. The image state at time $t$ is denoted as $\mathbf{X}(t)$, and its temporal evolution is governed by a parameterized vector field $\mathbf{f}$:

\begin{equation} \frac{d\mathbf{X}(t)}{dt} = \mathbf{f}(\mathbf{X}(t), \mathbf{Z}, t), \quad \mathbf{X}(0) = \mathbf{X}_0, \end{equation}

\noindent where $\mathbf{Z}$ denotes a task-specific prompt vector extracted from the input, which modulates the structure and intensity of the vector field. The vector field $\mathbf{f}$ determines both the direction and magnitude of the update, driving the degraded image state towards a clean reconstruction in a progressive manner.

To enable efficient numerical computation, we discretize the continuous dynamics using the explicit Euler scheme, leading to the following update rule:

\begin{equation} \mathbf{X}_{t+1} = \mathbf{X}_t + \Delta t \cdot \mathbf{f}(\mathbf{X}_t, \mathbf{Z}, t), \quad t = 0, \dots, T{-}1. \end{equation}

This formulation enables the integration of prompt guidance and dynamic evolution in a unified framework, facilitating adaptive restoration across varying degradation patterns.

\subsection{Prompt Generator}

The Prompt Generator is designed to encode degradation semantics from the input and produce a low-dimensional task prompt tensor $\mathbf{Z} \in \mathbb{R}^{B \times d \times 1 \times 1}$, which serves to enhance the restoration model's perceptual awareness and guide its reconstruction pathway. The generation process is defined as:

\begin{equation} \mathbf{Z} = 0.1 \cdot \sigma\left(\text{GN}_2\left(\text{Conv}_2\left(\text{GELU}\left(\text{GN}_1\left(\text{Conv}_1(\bar{\mathbf{X}})\right)\right)\right)\right)\right), \end{equation}

\noindent where $\bar{\mathbf{X}} = \text{mean}(\text{normalize}(\mathbf{X}))$ denotes the normalized and spatially averaged feature representation of the input frame. The output prompt $\mathbf{Z}$ is subsequently utilized for two purposes: (1) channel-wise modulation in the backbone network, and (2) semantic conditioning in the vector field evolution process.

This dual-purpose prompt design enables the model to adaptively adjust to varying degradation patterns while maintaining semantic consistency throughout the restoration trajectory.

\subsection{PhysicsUNet Network}

\textbf{PhysicsUNet} serves as the task-aware backbone for image restoration, designed with a symmetric U-Net architecture. Both the encoder and decoder modules are composed of unified \texttt{TaskBlock} units, which incorporate the prompt tensor $\mathbf{Z}$ via a residual gating mechanism to achieve feature modulation. The core computation is formulated as:

\begin{equation}
\mathbf{F}_{\text{mod}} = \text{ConvBlock}(\mathbf{X}) + \alpha \cdot \text{MLP}(\mathbf{Z}),
\end{equation}

\noindent where $\text{ConvBlock}(\mathbf{X})$ extracts convolutional features from the current image state $\mathbf{X}$, and $\text{MLP}(\mathbf{Z})$ projects the low-dimensional prompt tensor into channel-wise modulation factors, which are broadcasted across spatial dimensions and injected into the feature map. The scalar $\alpha$ is a learnable scaling coefficient (initialized as 0.1), which controls the strength of the prompt influence.

The resulting modulation feature $\mathbf{F}_{\text{mod}}$ plays two key roles in the framework:

\begin{itemize}
  \item \textbf{Feature Modulator:} $\mathbf{F}_{\text{mod}}$ preserves the spatial structure of the original image while injecting semantic priors from the prompt, acting as a task-aware feature representation that guides the subsequent restoration process.
  \item \textbf{Potential Constructor:} Rather than being directly used for output reconstruction, $\mathbf{F}_{\text{mod}}$ is passed into the potential computation module $\text{U}(\mathbf{X})$, contributing to the construction of the system’s Hamiltonian energy $\mathcal{H}(\mathbf{X}, \mathbf{P})$:
  \begin{equation}
  \mathbf{U}(\mathbf{X}) = \frac{1}{2} \left\| \mathbf{X} - \mathbf{x}_i \right\|^2,
  \end{equation}
  \begin{equation}
  \mathcal{H}(\mathbf{X}, \mathbf{P}) = \frac{1}{2} \cdot \mathbb{E}[\|\mathbf{P}\|^2] + U(\mathbf{X}),
  \end{equation}
  where $\mathbf{P}$ is the momentum tensor that characterizes the directional tendency of the image’s evolution, and $U(\mathbf{X})$ is derived by aggregating multi-scale $\mathbf{F}_{\text{mod}}$ features, reflecting the ``restoration potential energy'' of the current state. This energy term is later injected into the vector field calculation to guide the system toward lower-energy (i.e., clearer) image states, directly influencing the definition and behavior of the dynamic vector field.
\end{itemize}

During the decoding stage, modulation features from multiple layers are progressively fused with upsampled features via skip connections, eventually producing an image of the same resolution as the input. The final output of PhysicsUNet is denoted as $\tilde{\mathbf{X}}$:

\begin{equation}
\tilde{\mathbf{X}} = \text{PhysicsUNet}(\mathbf{X}_{\text{in}}, \mathbf{Z}),
\end{equation}

\noindent where $\mathbf{X}_{\text{in}}$ is the input degraded frame, and $\mathbf{Z}$ is the task prompt tensor generated by the Prompt Generator.

\subsection{Prompt-Guided Vector Field and ODE Solving}

The vector field $\mathbf{f}$ serves as the core driving force of image state evolution, integrating momentum, physical potential gradients, and prompt-based control terms. We first define the momentum tensor as:

\begin{equation}
\mathbf{P} = \tilde{\mathbf{X}} - \mathbf{X}_{\text{in}},
\end{equation}

\noindent where $\tilde{\mathbf{X}}$ is the initial restoration result from the backbone network, and $\mathbf{X}_{\text{in}}$ is the input degraded image. The momentum $\mathbf{P}$ characterizes the update direction of the image state and is treated as a constant inertial term during the evolution process.

The complete form of the vector field is given by:

\begin{equation}
\mathbf{f}(\mathbf{X}_t, \mathbf{Z}, t) = \frac{d\mathbf{P}}{dt} + \exp(-\lambda t) \cdot \tanh\left(-\frac{\partial \mathcal{H}}{\partial \mathbf{X}_t}\right) + \phi(\mathbf{Z}).
\end{equation}

The constructed vector field comprises three components: the first term captures the inertia effect induced by the momentum derivative; the second term introduces a physically-inspired decay mechanism, where the gradient of the Hamiltonian potential \( \frac{\partial \mathcal{H}}{\partial \mathbf{X}_t} \) is modulated by an exponential decay factor \( \exp(-\lambda t) \) to ensure stability; and the third term \( \phi(\mathbf{Z}) \) serves as a prompt control signal, mapping the task-specific prompt \( \mathbf{Z} \) into a spatial guidance field that directs the restoration process.

The entire evolution process is numerically solved using Euler integration, progressively updating the image state $\mathbf{X}_t$ over time. The final state $\mathbf{X}_T$ is taken as the restored output image.

\subsection{Loss Function}

We adopt the $\ell_1$ loss as the supervision signal to measure the pixel-wise discrepancy between the predicted image and the ground-truth clean image:

\begin{equation}
L = \| \mathbf{X}_{\text{pred}} - \mathbf{X}_{\text{gt}} \|_1,
\end{equation}

\noindent where $\mathbf{X}_{\text{pred}} = \mathbf{X}_T$ denotes the final restored image obtained at time step $T$, and $\mathbf{X}_{\text{gt}}$ is the corresponding ground-truth clean image. The $\ell_1$ loss is known to better preserve image edges and fine details compared to $\ell_2$ loss, thereby contributing to improved perceptual quality in restoration results.
\section{Experiments}
\subsection{Setup}
We evaluated the UniFlowRestore framework using multiple public video datasets, including YouTube-8M with millions of videos and UCF101 with 101 action classes. These datasets cover a wide range of video types, including daily life, sports, natural scenery, and traffic surveillance, under various conditions. To simulate real-world degradations, we introduced Gaussian and motion blurs (30\%), Gaussian and salt-and-pepper noise (25\%), compression artifacts (20\%), weather-related degradations (15\%), and others like scratches and occlusions (10\%). Experiments were conducted on single-card NVIDIA L20 and H20 GPUs, using PyTorch, Python 3.8, and libraries such as NumPy and SciPy. CUDA 11.3 and cuDNN 8.2 optimized GPU computations. For model training, we set the learning rate (\(lr\)) to \(1 \times 10^{-4}\), the batch size to 8, and trained the model for 500 epochs using 256$\times$256 resolution input frames (3-10 clips).

\subsection{Datasets}
For comprehensive evaluation of our video restoration algorithm across various degradation scenarios, we curated a unified all-in-one dataset, combining multiple representative video datasets covering tasks such as dehazing, deraining, denoising, and deblurring.

\noindent \textbf{Dehazing: RESIDE Dataset\cite{li2019benchmarking}.}
The RESIDE dataset, designed for single-image dehazing, consists of both synthetic and real-world hazy images. It includes diverse subsets for training (Indoor and Outdoor), objective testing (synthetic objective testing set), real-world task-driven testing (RTTS), and subjective testing with human scores (HSTS).

\noindent \textbf{Deraining: NTURain Dataset\cite{chen2018robust}.}
NTURain contains both synthetic and real-world rain-affected video frames. The dataset presents varied rainfall patterns, allowing for rigorous evaluation of deraining algorithms in complex real-world scenarios.

\noindent \textbf{Denoising: DAVIS (with synthetic noise) Dataset\cite{Pont-Tuset_arXiv_2017}.}
The DAVIS dataset, originally for video object segmentation, was extended for denoising by adding synthetic noise. It includes video sequences from both the 2016 and 2017 versions, offering a challenging benchmark for denoising algorithms.

\noindent \textbf{Deblurring: GoPro Dataset\cite{Nah_2017_CVPR}.}
The GoPro dataset, specifically designed for deblurring, includes 3,214 blurred images captured at 240fps. These images simulate motion blur using multiple frames, offering a realistic evaluation of deblurring techniques.

To boost the generality of our evaluation, we combined these datasets into a unified all-in-one video dataset. For each video, we performed frame extraction, randomly selecting 3 to 10 frames per second, ensuring a diverse mix of both real-world and synthetic scenarios. For the acquired video clips, the video segments are read randomly. In order to boost the generalization ability of the model, we also employ some data augmentation methods such as random cropping and rotation.
To ensure the fairness of the experiment, for all evaluation methods, these data augmentation strategies are applied when they are fine-tuned on our dataset until the loss converges stably.
\begin{figure*}[ht]
    \centering
    \includegraphics[width=0.98\linewidth]{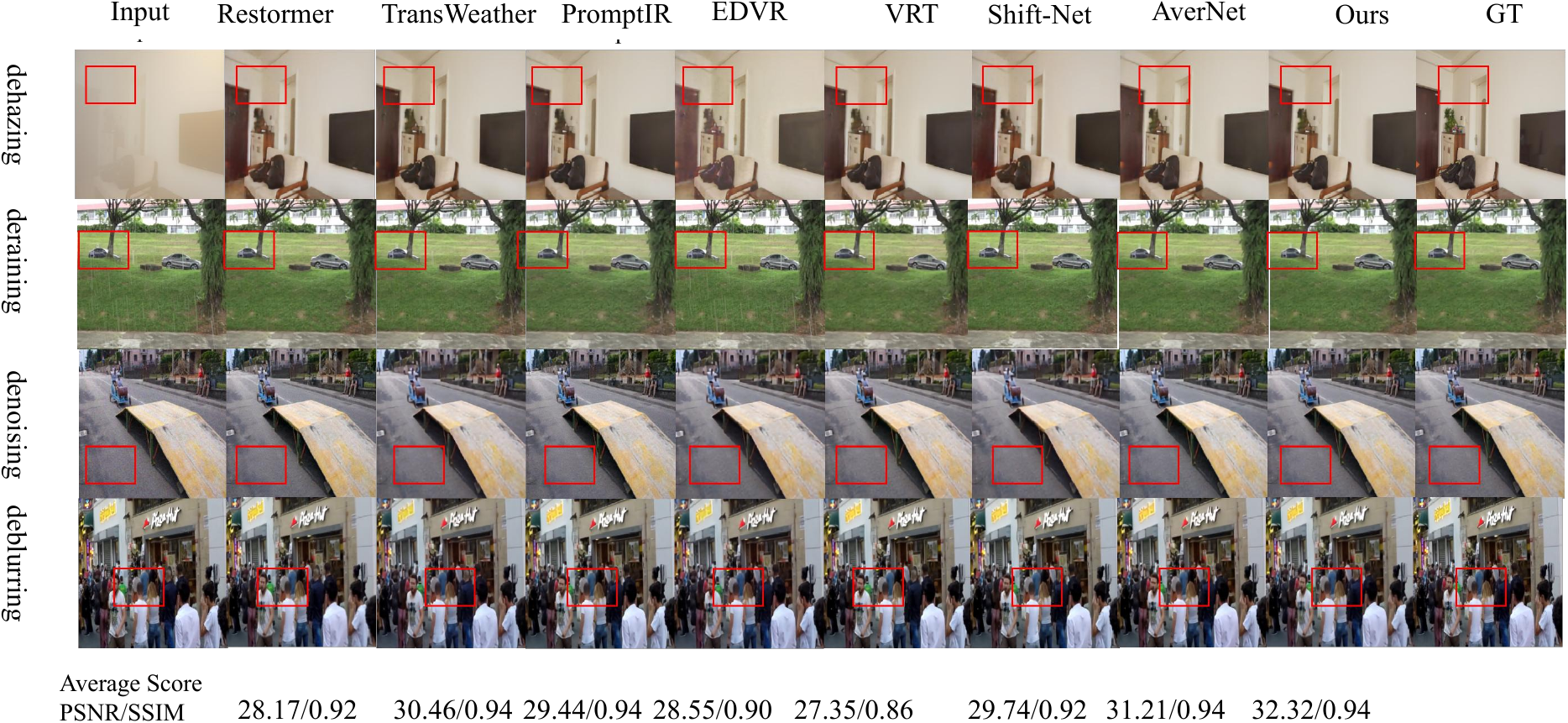}
    \caption{Visualization of all-in-one task in different models. Our approach reached state-of-art performance in most tasks.}
    \label{fig:flow_path}
    \vspace{-3mm}
\end{figure*}

\subsection{Qualitative and Quantitative Evaluation}
In this section, we present a comprehensive evaluation of the UniFlowRestore framework across multiple video restoration tasks, including dehazing, deraining, denoising, and deblurring. The performance of UniFlowRestore is compared with several state-of-the-art methods, and the results are summarized in Table~\ref{tab1}. We assess the framework based on Peak Signal-to-Noise Ratio (PSNR) and Structural Similarity Index (SSIM), two commonly used metrics for image and video restoration tasks.

\noindent \textbf{Dehazing Task.}
UniFlowRestore outperforms all other methods in terms of PSNR, achieving a peak value of 32.04dB, which is significantly higher than that of Restormer (28.78dB) and EDVR (27.05dB). While it slightly trails AverNet in SSIM, its performance in PSNR and the balanced SSIM value (0.94) make it a strong contender in this task. Notably, our method shows clear advantages over TransWeather, which achieves an SSIM of 0.89 and a PSNR of 31.41. These results suggest that UniFlowRestore is particularly effective at enhancing the visibility of objects and preserving fine details in hazy images.

\noindent \textbf{Deraining Task.}
For the deraining task, UniFlowRestore demonstrates superior performance across both PSNR and SSIM, with values of 31.90dB and 0.94, respectively. This is an improvement over Restormer (PSNR: 29.10dB, SSIM: 0.93) and Shift-Net (PSNR: 27.51dB, SSIM: 0.89). The effectiveness of UniFlowRestore in removing rain streaks while preserving important visual details is evident from the high PSNR and SSIM values. The results highlight that UniFlowRestore provides a cleaner, artifact-free appearance in rain-affected video frames.

\noindent \textbf{Denoising Task.}
In the denoising task, UniFlowRestore significantly outperforms other methods with a PSNR of 33.89dB and an SSIM of 0.97. The PSNR value is notably higher than that of other state-of-the-art methods, including Shift-Net (31.81dB), PromptIR (31.12dB), and AverNet (31.65dB). Moreover, the SSIM value (0.97) indicates that UniFlowRestore excels at maintaining structural integrity and texture while effectively reducing noise in video frames. This makes it particularly suitable for applications that demand both high-quality restoration and minimal loss of image features.

\noindent \textbf{Deblurring Task.}
UniFlowRestore achieves competitive results in the deblurring task, with a PSNR of 31.44dB and an SSIM of 0.93. Although it slightly trails AverNet in PSNR (32.98dB), its SSIM score is superior, indicating that it preserves image structure more effectively than the other methods. 

Overall, UniFlowRestore demonstrates state-of-the-art performance in all video restoration tasks, consistently outperforming or matching the best methods across multiple metrics. The framework shows a promising capability for handling diverse degradation types, offering a balanced trade-off between restoration quality and computational efficiency. 

\begin{table*}[htbp]
    \centering
    \caption{PSNR and SSIM Comparison of Different Methods for Each Task. The best performance are \textbf{bold} and the second are \underline{underlined}. Our method achieved the best performance in terms of the overall effect (average performance).}
    \begin{tabularx}{\textwidth}{l *{10}{X}} 
        \toprule
        {Method} & \multicolumn{2}{c}{Dehazing Task} & \multicolumn{2}{c}{Deraining Task} & \multicolumn{2}{c}{Denoising Task} & \multicolumn{2}{c}{Deblurring Task} & \multicolumn{2}{c}{Average Score} \\
        \cmidrule(lr){2 - 3} \cmidrule(lr){4 - 5} \cmidrule(lr){6 - 7} \cmidrule(lr){8 - 9}
        \cmidrule(lr){10 - 11} 
        
         & PSNR & SSIM & PSNR & SSIM & PSNR & SSIM & PSNR & SSIM & PSNR & SSIM \\
        \midrule
        Restormer\cite{zamir2022restormer} & 28.78 & 0.92 & 29.10 & \underline{0.93} & 26.72 & 0.90 & 28.06 & 0.92 & 28.67 & 0.92 \\
        TransWeather\cite{valanarasu2022transweather} & \underline{31.41} & 0.94 & \underline{30.43} & 0.93 & 30.88 & \underline{0.96} & 29.11 & 0.90 & 30.21 & \underline{0.94} \\
        PromptIR\cite{li2024promptcir} & 30.87 & \textbf{0.95} & 28.65 & \textbf{0.94} & 31.12 & 0.94 & 27.12 & 0.88 & 29.94 & 0.93 \\
        EDVR\cite{wang2019edvr} & 27.05 & 0.86 & 26.53 & 0.88 & 28.01 & 0.86 & 27.82 & 0.86 & 27.35 & 0.86 \\
        VRT\cite{liang2024vrt} & 27.96 & 0.89 & 28.58 & 0.86 & 29.63 & 0.93 & 28.02 & 0.92 & 28.30 & 0.90 \\
        Shift-Net\cite{li2023simple} & 28.58 & 0.89 & 27.51 & 0.89 & \underline{31.81} & 0.94 & 31.05 & \underline{0.94} & 29.99 & 0.91 \\
        AverNet\cite{zhao2024avernet} & 30.98 & \underline{0.94} & 29.23 & \underline{0.93} & 31.65 & 0.94 & \textbf{32.98} & \textbf{0.96} & \underline{31.22} & \underline{0.94} \\
        UniFlowRestore & \textbf{32.04} & \underline{0.94} & \textbf{31.90} & \textbf{0.94} & \textbf{33.89} & \textbf{0.97} & \underline{31.44} & 0.93 & \textbf{32.07} & \textbf{0.95} \\
        \bottomrule
    \end{tabularx}
    \vspace{-3mm}
    \label{tab1}
\end{table*}


\subsection{Ablation Study}
In this section, we conduct an ablation study to evaluate the influence of various components within the UniFlowRestore framework. Specifically, we compare the performance of the complete model, which incorporates Hamiltonian dynamics, with a variant that omits this component, focusing solely on the simplified physics-based vector field.

\noindent \textbf{Effect of Hamiltonian Dynamics in Vector Field Modeling.}
To assess the contribution of the Hamiltonian system in guiding the video restoration process, we contrast the full UniFlowRestore framework, which integrates the Hamiltonian energy formulation, with a simplified version lacking this dynamic system. The Hamiltonian version utilizes both momentum and potential energy to steer the restoration flow, as described by the equation:

\begin{equation}
f(X_t, Z, t) = \frac{dP}{dt} + \exp(-\lambda t) \cdot \tanh\left(-\frac{\partial H}{\partial X_t}\right) + \phi(Z).
\end{equation}

Conversely, the simplified model excludes the Hamiltonian energy term and relies exclusively on a physics-driven vector field without momentum. As illustrated in Table \ref{tab:ablations}, the comparison of restoration quality highlights the enhanced stability and superior quality achieved by the Hamiltonian-based system. Notably, it excels in preserving fine details and enabling smoother temporal transitions during multi-degradation video restoration tasks.

\noindent \textbf{Impact of Momentum Term.}
The momentum term \(P = \widetilde{X} - X_{in}\) plays a pivotal role in determining the direction of image state evolution. To explore its significance, we compare the full model with the momentum term included against a version where the momentum term is removed. In the ablation experiment without the momentum term, the system converges more slowly, and the restoration quality diminishes. The lack of inertia in the evolution process leads to more erratic updates. The momentum term facilitates maintaining coherence across consecutive video frames, thereby enhancing both spatial and temporal consistency.

\begin{figure}[t]
    \centering
    \includegraphics[width=1\linewidth]{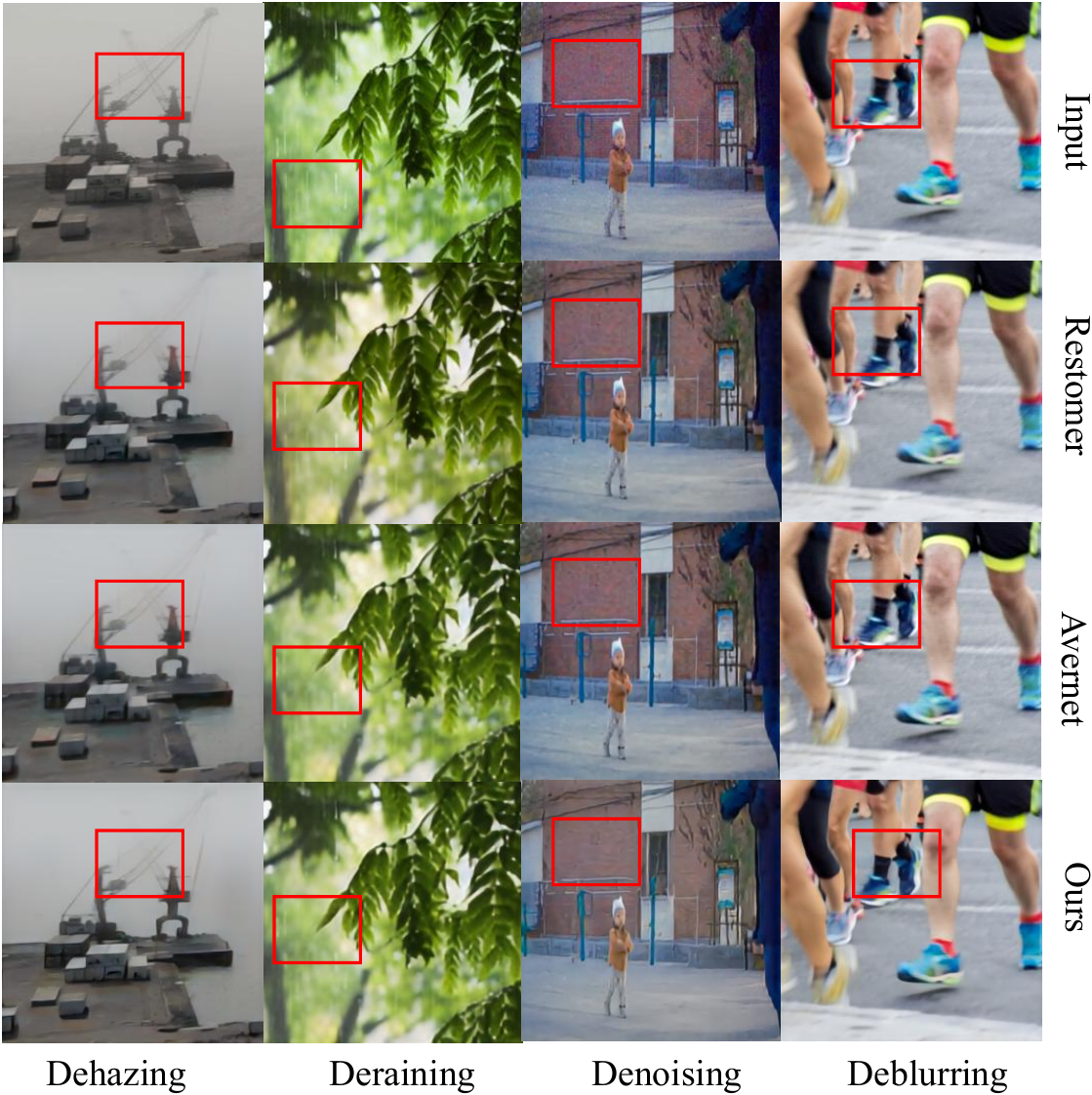}
    \caption{Qualitative results on real-world video frames with complex degradations. Top: degraded inputs. Bottom: restored outputs by UniFlowRestore. These methods were evaluated on these real-world images using the NIQE metric (the lower the score, the better). The average score of our method is 4.3, while the average scores of other methods range from 4.6 to 5.1.}
    \label{fig:realworld_vis}
    \vspace{-3mm}
\end{figure}

\begin{figure}[ht]
    \centering
    \includegraphics[width=1\linewidth]{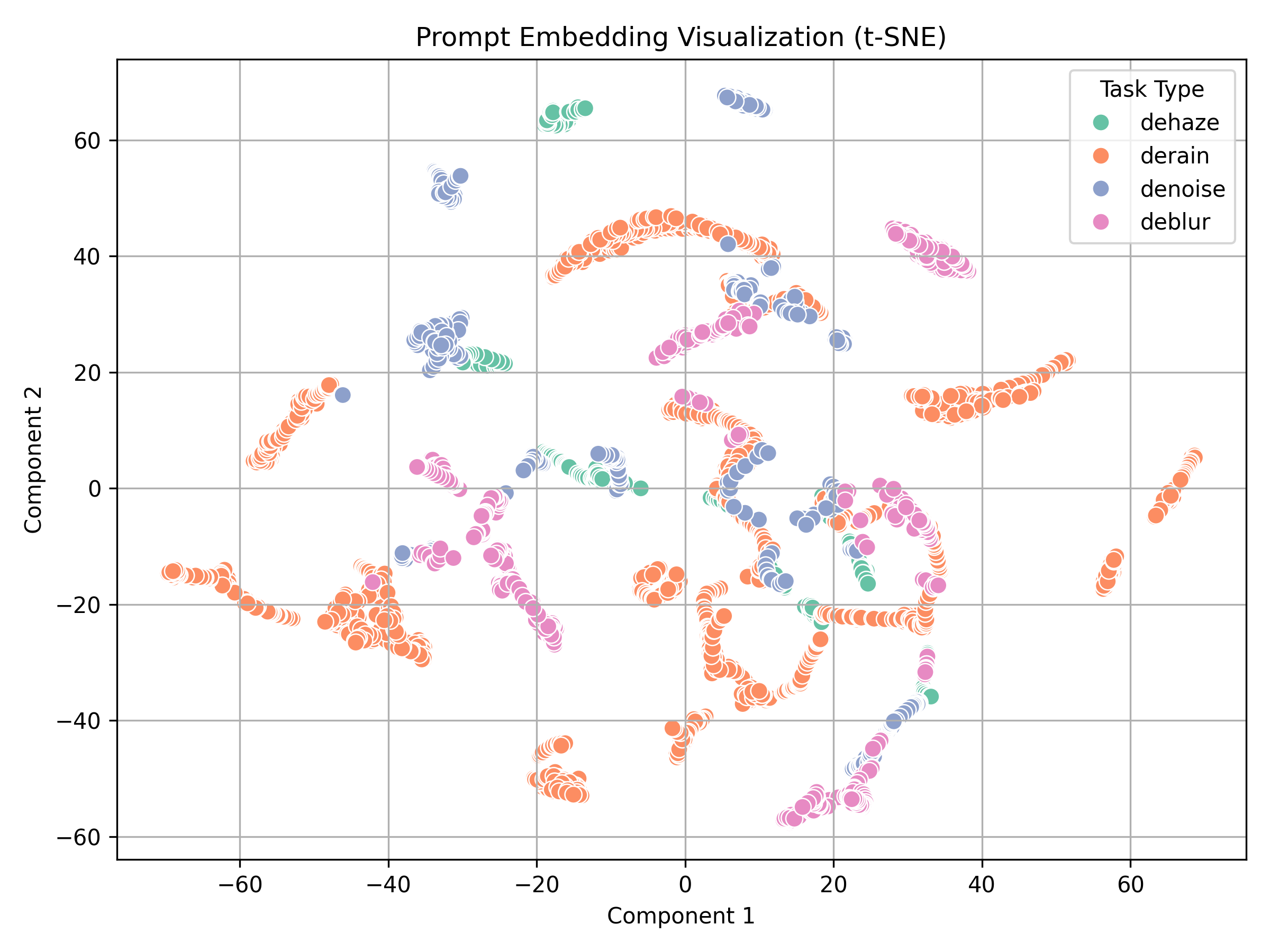}
    \caption{t-SNE visualization of prompt embeddings across four restoration tasks. Each point corresponds to a prompt vector extracted from one video frame. The clusters indicate that task semantics are effectively encoded by the PromptGenerator.}
    \label{fig:prompt_vis}
    \vspace{-3mm}
    
\end{figure}

\noindent \textbf{Role of Exponential Decay in Stability.}
We further analyze the contribution of the exponential decay term \(\exp(-\lambda t)\), which stabilizes the restoration process. Removing this decay term results in a model that may encounter instability during later restoration stages, giving rise to artifacts and over-corrections, especially in high-motion video scenes. Incorporating this decay factor ensures that the restoration process remains controlled and smooth over time, as evidenced by the reduction in artifacts and better preservation of visual details in the Hamiltonian version of the model.

\begin{figure*}[htbp]
    \centering
    \vspace{-2mm}
    \includegraphics[width=0.98\linewidth]{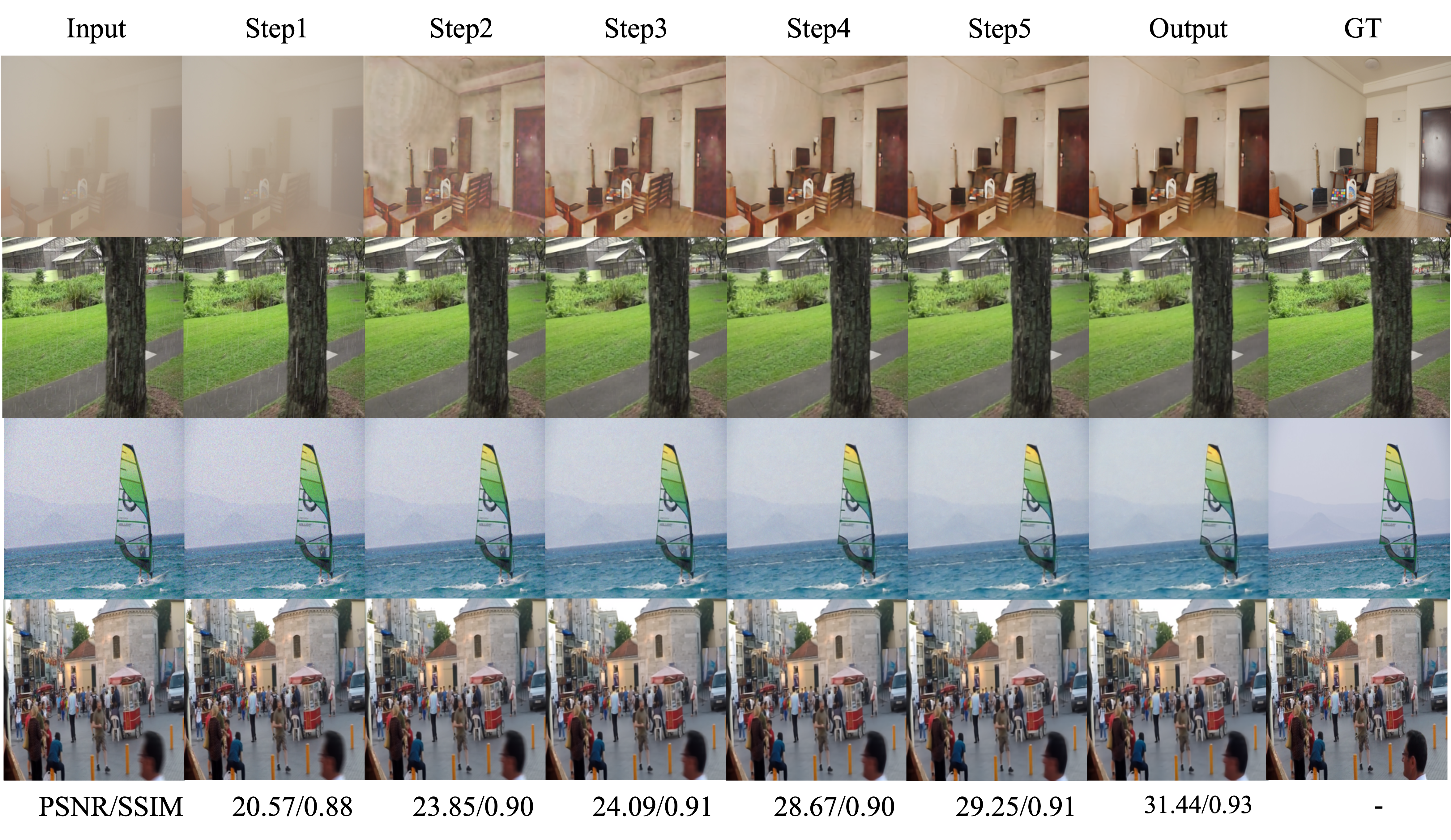}
    \vspace{-3mm}
    
    \caption{Visualization of intermediate results at each ODE step for four video restoration tasks. Each row shows the flow path from the degraded input to the final restored image. Our method progressively removes degradation and refines detail across steps. To provide a quantitative perspective, we additionally report the average PSNR and SSIM for each step, highlighting the consistent improvement in restoration quality.}
    \label{fig:flow_path}
    \vspace{-3mm}
    
\end{figure*}

\noindent \textbf{Comparison of Restoration Performance Across Tasks.}
Table \ref{tab:ablations} summarizes the quantitative results across various video restoration tasks, including dehazing, deraining, denoising, and deblurring. It compares the full Hamiltonian dynamics-based model with the simplified vector field model. We observe that the Hamiltonian system consistently outperforms the simpler model, particularly in scenarios involving multiple degradation types or dynamic degradation patterns over time.

\begin{table}[htbp]
\centering
\footnotesize
\caption{Quantitative Comparison of Performance Across Different Models in Various Video Restoration Tasks.}
\vspace{-3mm}
\begin{tabular}{lcccc}
\toprule
\multirow{2}{*}{\textbf{Task}} & \multicolumn{2}{c}{\textbf{Hamiltonian Model}} & \multicolumn{2}{c}{\textbf{Simplified Vector Field Model}} \\
\cmidrule(lr){2-3} \cmidrule(lr){4-5}
& \textbf{PSNR} & \textbf{SSIM} & \textbf{PSNR} & \textbf{SSIM} \\
\midrule
Dehazing   & 32.04 & 0.94 & 29.31 & 0.90 \\
Deraining  & 31.90 & 0.94 & 29.99 & 0.91 \\
Denoising  & 33.89 & 0.97 & 31.82 & 0.92 \\
Deblurring & 31.44 & 0.93 & 27.16 & 0.87 \\
\bottomrule
\end{tabular}
\vspace{-3mm}
\label{tab:ablations}
\end{table}

\section{Discussion}

This table clearly illustrates the advantages of incorporating Hamiltonian dynamics into the restoration process. The full model consistently outperforms the simplified variant in terms of both PSNR and SSIM, especially in the dehazing, deraining, and deblurring tasks, where the dynamics contribute to enhanced temporal coherence and finer details.

\noindent \textbf{Real-world video restoration.} To assess the generalization capability of our framework, we evaluate it on real-world video frames exhibiting diverse and complex degradations, including haze, rain streaks, sensor noise, and dynamic motion blur. Figure~\ref{fig:realworld_vis} presents qualitative results on four representative scenes, where the top row shows the degraded inputs and the bottom row shows the outputs restored by UniFlowRestore. Our method consistently enhances visibility while preserving structural fidelity and color consistency across varied conditions. In hazy and rainy scenes, it effectively removes low-contrast artifacts without over-smoothing or introducing halos. For scenes with motion blur and cluttered backgrounds, our model recovers fine details and maintains spatial coherence. These results demonstrate the strong robustness and real-world applicability of our unified restoration framework without relying on task-specific priors or tuning.

\noindent\textbf{Prompt response visualization.} As shown in Figure~\ref{fig:prompt_vis}, we visualize the learned prompt embeddings from four restoration tasks using t-SNE. Each point represents a prompt vector from a video frame, colored by task type. We observe that the embeddings from each task form distinct clusters in the 2D space, suggesting that our PromptGenerator effectively encodes task-specific semantics. In particular, prompt vectors from \textit{derain} and \textit{deblur} partially overlap, indicating similarity in their degradation patterns. These results show that the prompt module learns meaningful and compact representations that help the network adapt to diverse tasks without explicit task labels.

\noindent\textbf{Flow path visualization.} Figure~\ref{fig:flow_path} shows the restoration process of four degraded video frames over multiple ODE steps. Each row visualizes one frame undergoing five intermediate updates from the initial degraded input to the final output.
We find that in the early steps, the model focuses on removing dominant degradation (e.g., haze, rain, or noise). In later steps, it enhances fine textures, edges, and colors. This step-by-step evolution confirms that our model follows a continuous and interpretable restoration trajectory guided by prompt and physical priors.

\noindent\textbf{Complexity of the model.} Table~\ref{tab:complexity} summarizes the model complexity and runtime comparison. UniFlowRestore achieves the lowest computational cost among all compared methods, with only 1.15M parameters and 18.33G MACs. It also demonstrates the fastest runtime (0.09s per frame), making it well-suited for deployment in resource-constrained scenarios. Compared to VRT and AverNet, our model achieves up to 39× and 7× reduction in MACs, respectively.

\begin{table}[htbp]
\centering
\footnotesize
\caption{Comparison of model complexity and runtime efficiency.}
\vspace{-3mm}
\begin{tabular}{lccc}
\toprule
\textbf{Model} & \textbf{Params (M)} & \textbf{MACs (G)} & \textbf{Runtime (s)} \\
\midrule
VRT~\cite{liang2024vrt}           & 18.3   & 721.3   & 0.42 \\
Shift-Net~\cite{li2023simple}    & 4.1    & 47.1    & 0.12 \\
AverNet~\cite{zhao2024avernet}       & 6.5    & 135.0   & 0.24 \\
\textbf{UniFlowRestore (Ours)} & \textbf{1.15}  & \textbf{18.33}  & \textbf{0.09} \\
\bottomrule
\end{tabular}
\vspace{-6mm}
\label{tab:complexity}
\end{table}

\noindent\textbf{Limitations and future work.} While UniFlowRestore demonstrates strong generalization across diverse video restoration tasks, there are still some limitations. First, although the proposed framework supports unified restoration without task-specific annotations, its performance can slightly degrade under extreme degradations not represented in the training data. Second, the current model design assumes a relatively consistent degradation level within each video clip; abrupt temporal degradation shifts may challenge the stability of the vector field.  Future work will explore adaptive step solvers and degradation-aware modulation strategies to further enhance robustness and scalability. In addition, the quality of the dataset is also a crucial issue. In our follow-up work, we will introduce techniques such as video low-light enhancement and video HDR (High Dynamic Range) enhancement.

\section{Conclusion}

In this paper, we present UniFlowRestore, a unified video restoration framework based on flow matching and Prompt Guidance. The proposed method integrates a physics-inspired evolution mechanism with task-adaptive prompt control to address multiple types of video degradation in a single model. From the perspective of interpretability, our approach gradually removes dominant degradation in early steps and progressively refines structural and color information through a continuous vector field. Extensive experiments show that UniFlowRestore achieves a strong generalization across diverse video restoration tasks.

\bibliographystyle{ACM-Reference-Format}
\bibliography{sample-base}

\end{document}